%% file: arXiv.tex
\documentclass[10pt,twocolumn,letterpaper]{article}

\usepackage[pagenumbers]{iccv} 

\input{preamble}

\definecolor{iccvblue}{rgb}{0.21,0.49,0.74}
\usepackage[pagebackref,breaklinks,colorlinks,allcolors=iccvblue]{hyperref}
\hypersetup{
    colorlinks=true,
    urlcolor=magenta  
}

\def\paperID{6728} 
\def\confName{ICCV}
\def\confYear{2025}

\title{Progressive Growing of Video Tokenizers for Temporally Compact Latent Spaces}
\author{
  Aniruddha Mahapatra$^{1}$\ \ \
  Long Mai$^{1}$\ \ \ 
  David Bourgin$^{1}$ \ \ \
  Yitian Zhang$^{1,2}$\textsuperscript \ \ \ 
  Feng Liu$^{1}$\\[1ex]
    $^{1}$Adobe Research\ \ \ 
    $^{2}$Northeastern University\\[1ex]
\url{https://progressive-video-tokenizer.github.io/Pro-MAG/}
}

\begin{document}

\input{macros/common}
\input{macros/authors}

\input{figures/teaser}



\begin{abstract}
Video tokenizers are essential for latent video diffusion models, converting raw video data into spatiotemporally compressed latent spaces for efficient training. However, extending state-of-the-art video tokenizers to achieve a temporal compression ratio beyond 4× without increasing channel capacity poses significant challenges.
In this work, we propose an alternative approach to enhance temporal compression. We find that the reconstruction quality of temporally subsampled videos from a low-compression encoder surpasses that of high-compression encoders applied to original videos. This indicates that high-compression models can leverage representations from lower-compression models.
Building on this insight, we develop a bootstrapped high-temporal-compression model that progressively trains high-compression blocks atop well-trained lower-compression models. Our method includes a cross-level feature-mixing module to retain information from the pretrained low-compression model and guide higher-compression blocks to capture the remaining details from the full video sequence.
Evaluation of video benchmarks shows that our method significantly improves reconstruction quality while increasing temporal compression compared to 
directly training the full model. Furthermore, the resulting compact latent space effectively trains a video diffusion model for high-quality video generation with a significantly reduced token budget.
\end{abstract}

\vspace{-0.2in}
\section{Introduction}
\label{sec:intro}

\input{Introduction}

\vspace{-0.05in}
\section{Related work}
\label{sec:relw}

\input{Related_Work}

\vspace{-0.1in}
\section{Method}
\label{sec:method}
\input{Method}

\section{Experiments}
\label{sec:exp}
\input{Experiments}

\section{Discussion}
\label{sec:disc}
\input{Discussion}

\vspace{-0.1in}
\section{Conclusion and Limitations}
\label{sec:concl}
\vspace{-0.1in}
\input{Conclusion}

\myparagraph{Acknowledgments.} We thank Kuldeep Kulkarni, Cusuh Ham, and Gunjan Aggarwal for proofreading the paper. We are also grateful to Richard Zhang, Zhifei Zhang, Jianming Zhang, Gunjan Aggarwal, Kevin Duarte, Yang Zhou, Zhan Xu, and Lakshya for several fruitful discussions. This work was partly done while Yitian was an intern at Adobe.

{
    \small
    \bibliographystyle{ieeenat_fullname}
    \bibliography{main}
}

\appendix
\input{appendix}


\end{document}

%% file: preamble.tex
\usepackage[ruled]{algorithm2e} 

\SetAlFnt{\small}
\SetAlCapFnt{\small}
\SetAlCapNameFnt{\small}
\SetAlCapHSkip{0pt}

\definecolor{cvprblue}{rgb}{0.21,0.49,0.74}

\usepackage{graphicx}
\usepackage{amsmath}
\usepackage{amssymb}
\usepackage{booktabs}

\usepackage{color}
\usepackage{makecell}

\definecolor{tabhigh}{rgb}{0.9, 0.25, 0.25}
\definecolor{tabmid}{rgb}{0.9, 0.6, 0.05}
\definecolor{tablow}{rgb}{0.4, 0.8, 0.35}

\usepackage{tikz}
\usepackage{pgfplots}
\pgfplotsset{compat=1.18}
\usepackage{graphicx}
\usepackage{amsmath}
\usepackage{amssymb}
\usepackage{booktabs}
\usepackage{url}
\usepackage{epsfig}
\usepackage{lipsum}
\usepackage{amsthm}
\usepackage{comment}
\usepackage{multirow}
\usepackage{subcaption}
\usepackage{microtype}
\usepackage{xspace}
\usepackage{enumitem}
\usepackage{graphbox}
\usepackage{epigraph}
\usepackage{wrapfig}
\usepackage{textcomp}  
\usepackage{scalerel}  
\usepackage{animate}
\usepackage{colortbl}
\usepackage{tabularx,booktabs}

\usepackage{multirow}
\usepackage{dsfont}
\usepackage{enumitem}

\usepackage{algorithm}
\usepackage{algpseudocode}
\usepackage{animate}
\usepackage{booktabs} 
\usepackage{hhline}
\usepackage{adjustbox}
\usepackage{multirow}
\usepackage{booktabs}
\usepackage[normalem]{ulem}
\usepackage{ifthen}
\usepackage{epsfig}
\usepackage{graphics}
\usepackage{graphicx}
\usepackage{amsmath}
\usepackage{animate}
\usepackage{amsmath}
\usepackage{booktabs}
\usepackage{hhline}
\usepackage{adjustbox}
\usepackage{booktabs,arydshln}
\usepackage{amsfonts}
\usepackage{color}

%% file: macros/common.tex
\newcommand{\reffig}[1]{Figure~\ref{fig:#1}}
\newcommand{\refsec}[1]{Section~\ref{sec:#1}}
\newcommand{\refapp}[1]{Appendix~\ref{sec:#1}}
\newcommand{\reftbl}[1]{Table~\ref{tab:#1}}
\newcommand{\refalg}[1]{Algorithm~\ref{alg:#1}}
\newcommand{\refline}[1]{Line~\ref{line:#1}}
\newcommand{\shortrefsec}[1]{\S~\ref{sec:#1}}
\newcommand{\refeq}[1]{Equation~\ref{eq:#1}}
\newcommand{\refeqshort}[1]{(\ref{eq:#1})}
\newcommand{\shortrefeq}[1]{\ref{eq:#1}}
\newcommand{\lblfig}[1]{\label{fig:#1}}
\newcommand{\lblsec}[1]{\label{sec:#1}}
\newcommand{\lbleq}[1]{\label{eq:#1}}
\newcommand{\lbltbl}[1]{\label{tab:#1}}
\newcommand{\lblalg}[1]{\label{alg:#1}}
\newcommand{\lblline}[1]{\label{line:#1}}

\definecolor{MyDarkBlue}{rgb}{0,0.08,1}
\newcommand{\camera}[1]{#1}

\newcommand{\myparagraph}[1]{\vspace{-4pt}\paragraph{#1}}
\newcommand{\feature}{\bs{f}_{t}\xspace}



\newcommand{\bs}[1]{{\boldsymbol{#1}}}
\newcommand{\pixel}{p\xspace} 
\newcommand{\art}{\bs{x}\xspace} 
\newcommand{\nat}{\hat{\bs{x}}\xspace} 
\newcommand{\artt}{\bs{x}_t\xspace} 
\newcommand{\natt}{\hat{\bs{x}}_t\xspace}  
\newcommand{\arttminus}{\bs{x}_{t-1}\xspace}  
\newcommand{\nattminus}{\hat{\bs{x}}_{t-1}\xspace}  
\newcommand{\atten}{\bs{A}_t\xspace}  
\newcommand{\attenavg}{\bs{\overline{A}}\xspace}  
\newcommand{\artmask}{\bs{M}\xspace}   
\newcommand{\natmask}{\hat{\bs{M}}\xspace}  
\newcommand{\artflow}{\bs{F}\xspace}  
\newcommand{\natflow}{\hat{\bs{F}}\xspace} 

\newcommand{\at}{\bs{c}\xspace}   
\newcommand{\nt}{\hat{\bs{c}}\xspace} 
\newcommand{\gf}{\bs{G}_{flow}\xspace}
\newcommand{\gv}{\bs{G}_{frame}\xspace}

\newcommand{\magvit}{\text{MagViT-v2}\xspace}
\newcommand{\lio}{\text{ProMAG}\xspace}
\newcommand{\fourx}{4$\times$\xspace}
\newcommand{\eightx}{8$\times$\xspace}
\newcommand{\sixteenx}{16$\times$\xspace}




%% file: macros/authors.tex
\definecolor{myblue}{rgb}{0.239,0.553,0.565}

\definecolor{cmu_red}{rgb}{0.706,0.169,0.212}

%% file: figures/teaser.tex
\twocolumn[{
\renewcommand\twocolumn[1][]{#1}%
\maketitle

\vspace{-13mm}
\begin{center}
    \centering 
    \includegraphics[width=0.99\linewidth]{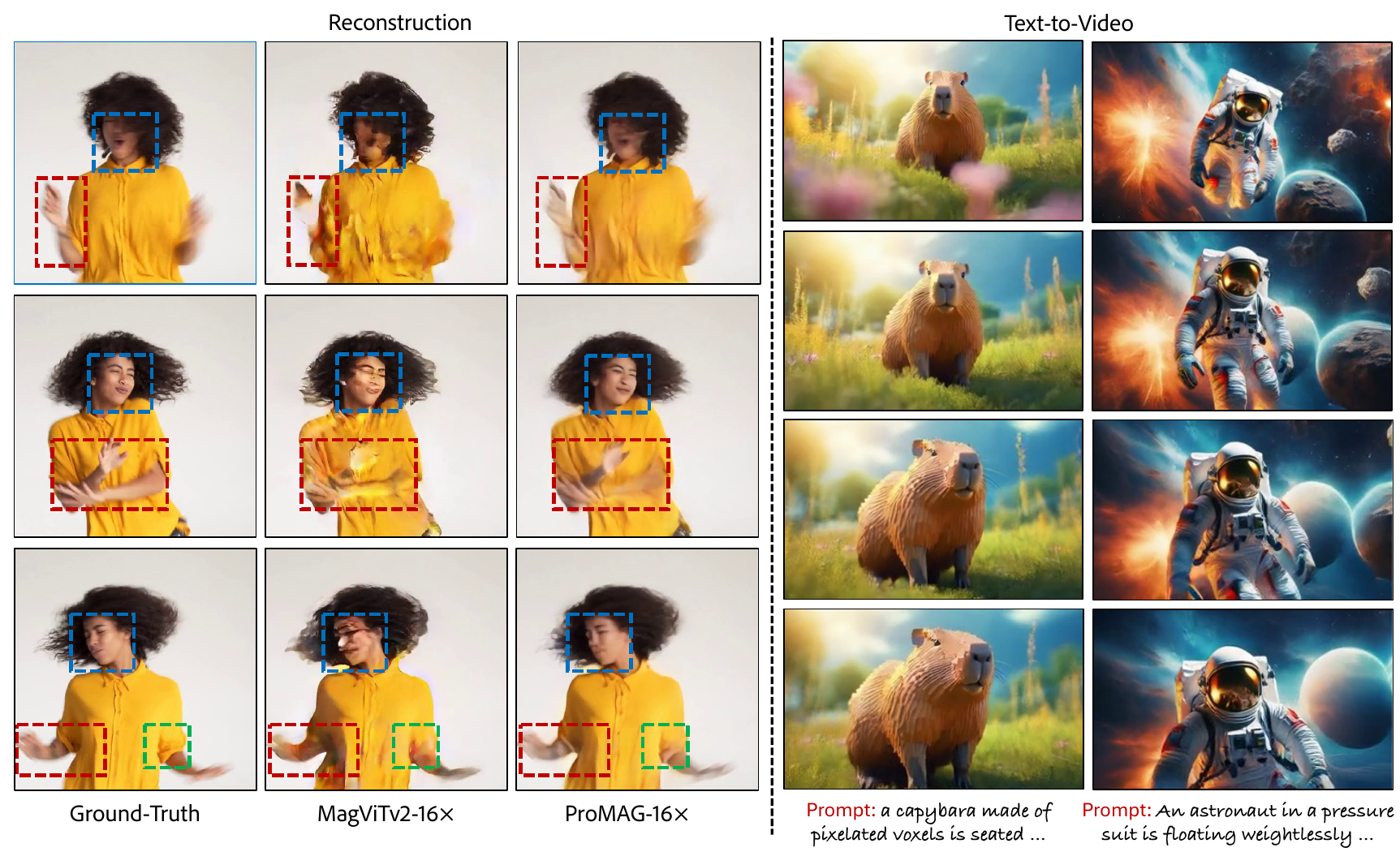} \vspace{-3mm}
    \captionof{figure}{ 
(left) Our video tokenizer \lio is capable of much more effectively reconstructing high motion videos even at very large (\sixteenx) temporal compression, compared to baseline \magvit. Dotted boxes highlight regions where the \magvit fails catastrophically, like faces and hands, which have a lot of artifacts in the reconstructions. (right) We show that our highly compressed latent space (\sixteenx) is suitable for training high-quality text-to-video diffusion models.
    } 
    \vspace{-2.5mm}
    \lblfig{teaser}
\end{center}

}]

%% file: Introduction.tex
The emergence of diffusion models~\cite{ho2020ddpm,song2021ddim} has transformed image~\cite{saharia2022photorealistic,ramesh2022hierarchical,gafni2022make,rombach2022high,peebles2023scalable} and video~\cite{ge2023preserve,blattmann2023videoldm,videocrater,singer2022make,ho2022imagen,yang2024cogvideox,videoworldsimulators2024} generation. These models gradually transform random noise into coherent visual outputs, showcasing impressive capabilities in producing high-fidelity content \cite{videoworldsimulators2024,esser2024scaling}.
Latent diffusion models (LDMs)\cite{rombach2022high} have gained popularity for image and video generation\cite{saharia2022photorealistic,balaji2022ediff}. Unlike pixel diffusion models that operate directly on raw pixels, LDMs project pixels into a low-dimensional latent space using a variational autoencoder (VAE)~\cite{kingma2013auto}, enabling diffusion in this compact space. This dimensionality reduction enhances computational efficiency, which is particularly crucial for video generation due to the higher dimensionality of video data compared to images.


Early latent video diffusion models (LVDMs) relied on latents extracted from each frame independently \cite{singer2022make, videocrater, blattmann2023stable} by reusing the same VAE from image LDMs. This neglects the temporal dynamics inherent in video data, compromising the temporal consistency of the generated videos, and offers no temporal compression in the latent space.
To address these limitations, the seminal work \magvit~\cite{yu2023language} pioneers a spatiotemporal video tokenizer that can jointly encode images and videos in the same latent space. Originally designed to encode video into discrete tokens, \magvit has been widely adapted to continuous tokens for use in many state-of-the-art LVDMs \cite{gupta2025photorealistic, yang2024cogvideox, opensora, pku_yuan_lab_and_tuzhan_ai_etc_2024_10948109}. In the context of this work, we use \magvit as a continuous tokenizer.

State-of-the-art video diffusion models (VDMs) utilize diffusion transformers~\cite{peebles2023scalable} as their backbone architectures, with computational costs scaling quadratically with input token lengths. Increasing latent space compression by a factor of two in both spatial and temporal dimensions significantly enhances model efficiency. While \magvit and other video tokenizers achieve substantial spatial compression (\eightx), their temporal compression remains limited to \fourx. This work focuses on improving temporal compression while maintaining spatial compression at \eightx.
We found that extending \magvit by adding more temporal down/upsampling layers to boost temporal compression negatively impacts reconstruction quality. This underscores a key challenge in adapting convolutional video tokenizers for higher compression rates: training such models from scratch necessitates simultaneous optimization of all parameters for high compression, rather than employing a hierarchical approach where different components specialize in varying levels of compression.

Interestingly, we observe that \fourx temporal compression \magvit can accurately reconstruct videos with \fourx frame subsampling (i.e., low FPS), much better than a \sixteenx temporal compression model on the original (high FPS) video. Thus, we try to answer the question: \textit{``Can we meaningfully boost the temporal compression of video tokenizers by reusing a pretrained \fourx model, \textbf{while keeping the number of latent channels constant}?''} 
In this paper, we intentionally keep the latent channel dimension fixed over the progressive growing process for two reasons. First, it helps separate the effects of channel dimensions and progressive growing on the model performance. Second, one of our main goals is to achieve good latent features not only for high reconstruction quality but also for diffusion model training. It has been observed that increasing the latent channel dimensions, while leading to a slight improvement in reconstruction, tends to make generative model training more difficult~\cite{yu2023language}. 

\noindent This paper makes the following main contributions:
\begin{itemize}
    \item We adopt the continuous-token \magvit architecture as our base model and make several modifications to make it more efficient and amenable to our method of growing the model from \fourx to \eightx and \sixteenx compression. The modifications preserve the quality of the base \fourx \magvit. We call our model \textbf{\lio} (abbreviation for \textbf{Pro}gressive growing of \textbf{Mag}ViT-v2 for continuous-space tokens).
    \item We present our progressive model growing technique to train \lio for high temporal compression ratios, achieving up to \eightx and even \sixteenx. Notably, we are the first to achieve high-quality reconstruction with a \sixteenx temporal compression model. Evaluations on video benchmarks show that \lio delivers significantly better reconstruction quality than simply extending existing tokenizers for higher compression. Our method also consumes $\sim2.7\times$ less training time compared to full model training.
    \item We showcase the effectiveness of our highly compressed latents for text-to-video generation with DiT. Experiments on the text-to-video evaluation benchmark~\cite{huang2024vbench} reveal that DiT trained with our \sixteenx temporal-compression latents achieves comparable or slightly higher quality than the standard \fourx compression, while significantly enhancing efficiency and token length.
\end{itemize}



%% file: Related_Work.tex
\noindent \textbf{Latent video diffusion models.} The pioneering LVDMs~\cite{blattmann2023stable, singer2022make, videocrater, ge2023preserve, blattmann2023align, menapace2024snap} repurposed image VAEs to transform each frame of the video independently into the latent space. This caused temporal flickering in the generated video and limited the video length due to the limited compression achieved by image VAEs. More recent works~\cite{gupta2025photorealistic, yang2024cogvideox, opensora, pku_yuan_lab_and_tuzhan_ai_etc_2024_10948109, videoworldsimulators2024} use VAEs that operate on video-level, hence mitigating both limitations of using image VAEs for video diffusion models. Building upon this, we aim to create a VAE with even higher temporal compression (\eightx and \sixteenx) compared to the contemporary \fourx, enabling highly efficient and longer video generation. 

\noindent \textbf{High compression VAEs.} There have been few works recently that investigate very high compression in latent space for higher-resolution image and longer video generation. UltraPixel~\cite{ren2024ultrapixel} provides an image autoencoder that can perform 24$\times$ spatial compression. Very recently, DC-AE~\cite{chen2024deep} provide a solution to increase the spatial compression to as large as 128$\times$ spatially. 
However, they also have to increase channel dimension consistently to preserve reconstruction quality. MovieGen~\cite{polyak2024movie} and Cosmos~\cite{agarwal2025cosmos} also achieve \eightx temporal compression by making modifications to the tokenizer architecture.
In comparison, we tackle a different problem in this work. Starting from a well-trained \fourx temporal compression model, our goal is to create a method to increase the temporal compression while keeping the number of latent channels fixed to avoid making the latent space harder to learn for the diffusion model. In this work, we keep the spatial compression fixed at \eightx.
To the best of our knowledge, we are also the first method to achieve \sixteenx temporal compression.

\input{figures/interpolation}

\noindent \textbf{Hierarchical models.} Hierarchical generation has been explored most notably using generative adversarial networks (GANs)~\cite{goodfellow2014generative} to perform high-resolution image generation in stages~\cite{wang2018pix2pixHD, karras2019style, karras2018progressive}. These approaches first learn a generator that operates in low resolution, followed by learning additional high-resolution blocks on top in stages. Similar ideas have also been explored in image and video generation using diffusion models~\cite{ren2024ultrapixel, balaji2022ediff, ho2022imagen, gafni2022make}, where content is first generated in low resolution and then upsampled in the spatial and temporal domain using a separate model. In our work, we are inspired by ProGAN~\cite{karras2018progressive} and adapt the progressive learning idea to our problem of boosting the temporal compression in video tokenizers, progressively learning separate model blocks to handle different levels of compression.

\input{figures/custom_groupnorm}

%% file: figures/interpolation.tex
\begin{figure}[t!]
    \centering
    \includegraphics[width=0.9\linewidth]{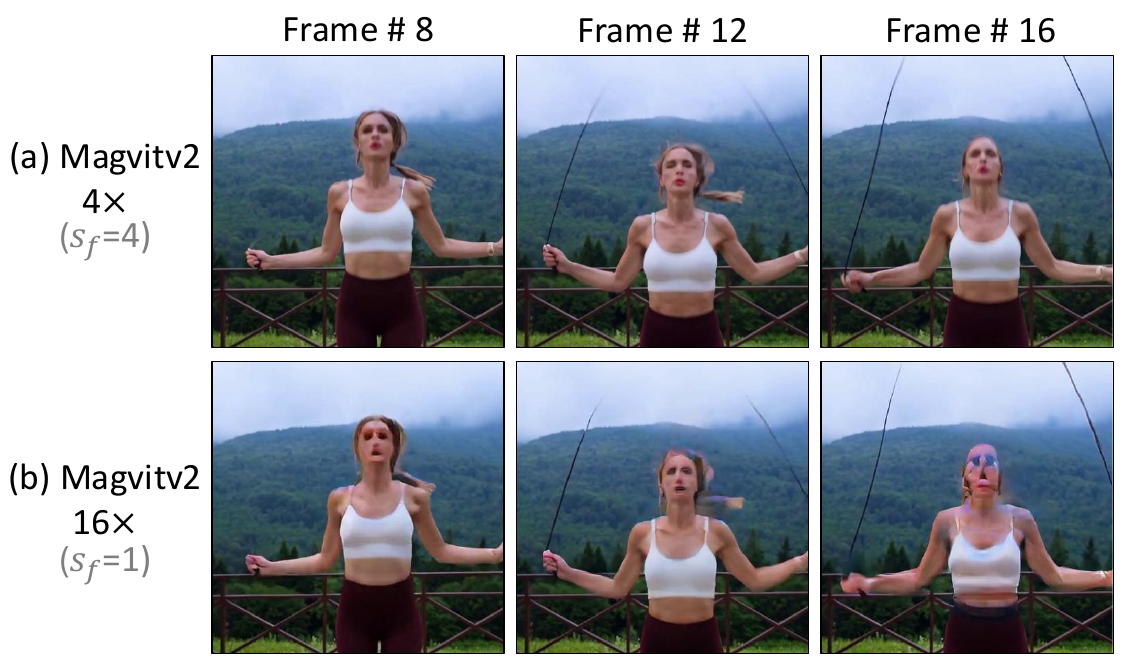}
    \vspace{-2mm}
    \caption{\camera{{\bf Motivation.} We highlight the key motivation of our progressive growing approach. Directly training \magvit for \sixteenx temporal compression leads to poor reconstruction quality for a $24$-fps video (bottom). $s_f$ stands for frame subsampling factor. However, we observed that the \fourx temporal compression model can still accurately reconstruct a $6$-fps video by feeding the same $24$-fps video after subsampling frames by a factor of $4\times$, $s_f=4$ (top). This observation implies that \textit{it is not necessarily the large motion that leads to worse reconstruction, but that training many downsampling (upsampling) layers of encoder (decoder) at once makes training difficult}, as also observed in DC-AE~\cite{chen2024deep}.
    }}
    \lblfig{interpolation}
\vspace{-4mm}
\end{figure}

%% file: figures/custom_groupnorm.tex
\begin{figure}[t!]
    \centering
    \includegraphics[width=0.9\linewidth]{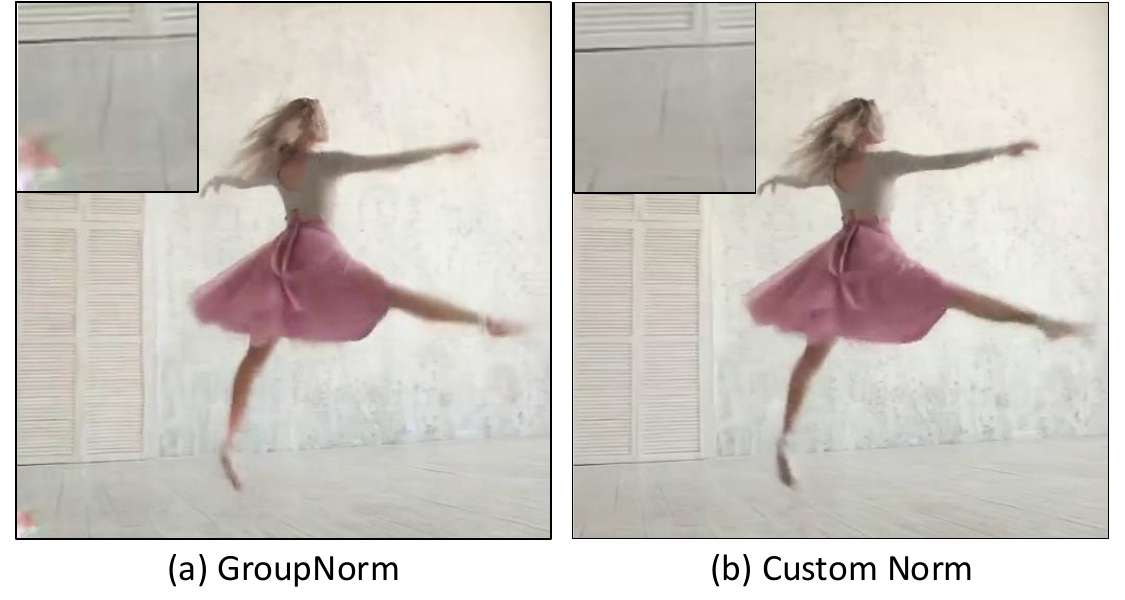}
    \vspace{-2mm}
    \caption{\textbf{`spot artifacts'. }{Our progressive growing approach with \texttt{GroupNorm} leads to the `spot' like artifacts (left) at the bottom right corner in reconstructed videos. (right) Removing the mean subtraction from group normalization eliminates the spot artifacts in reconstructed frames}.}
    \lblfig{custom_gn}
\vspace{-4mm}
\end{figure}

%% file: Method.tex
\input{figures/method}
Our goal is to create a continuous space video tokenizer with high temporal compression ratios (\eightx or \sixteenx) and good reconstruction quality. 
 In \refsec{base_magvitv2}, we discuss the modifications we make to the \magvit~\cite{yu2023language} to arrive at our  (\lio) base model. In \refsec{motivation}, we analyze why the base \lio, capable of doing \fourx temporal compression, has difficulties extending to \eightx or \sixteenx temporal compression. We discuss a method of progressively growing the \lio-\fourx model to achieve \eightx and consequently \sixteenx temporal compression in \refsec{pro_grow}. Finally, in \refsec{tile} we introduce our layer-wise spatial tiling technique during decoding to enable high-resolution encoding-decoding.

\subsection{Base model: \lio}
\label{sec:base_magvitv2}

Following prior works~\cite{qin2024xgen, yang2024cogvideox}, we build our model based on the continuous-token variant of \magvit~\cite{yu2023language}. The \magvit tokenizer is composed of 3D causal convolution layers. The use of causal 3D convolution not only enables full spatiotemporal processing but also facilitates the consistent encoding of images and video data. 
Our model encodes an input video (or image) with $1 + k \times N$ frames into $1 + \text{ N }$ latent frames, where $k$ is the compression ratio and $N=0$ when encoding images. \magvit, by default, can perform \fourx temporal compression ($k=4$) very reliably. 

We make three key modifications to the base \magvit model to increase training and inference efficiency and to make it amenable to our method of progressive growing. We call our model \lio.

\noindent\textbf{Image model initialization.} Instead of training a  2D image encoder and using the weights to initialize 3D convolution kernels as in \magvit~\cite{yu2023language}, we freeze the encoder of our pretrained 2D image model and use it to encode the first frame of the input video. We then train the \magvit model with 3D causal convolutions to focus only on encoding the rest of the video frames. In our experience, this makes the model training converge faster. 

\noindent\textbf{Efficient upsampling.} \magvit encodes a 17-frame video into $5$ latent frames which are decoded to $20$ frames after \fourx temporal upsampling. The presence of three additional frames in the reconstruction is a byproduct of the fact that the first latent frame only represents a single input frame. The \magvit decoder discards these $3$ reconstruction frames when generating the final output. We find this operation to be wasteful in terms of computation and becomes a greater problem with higher temporal compression. For example, in the case of \sixteenx compression on $17$ frames, the decoder will output $32$ frames and have to reject the first $15$ frames. To mitigate this issue, in all the temporal upsampling layers of the decoder, we discard the $1^{st}$ frame, as it can be considered a padding frame in causal convolutions. This reduces the memory consumption significantly, especially for higher compression models, without hurting the reconstruction quality. 

\noindent\textbf{Removing \textit{`spot'} artifacts.} \magvit uses group normalization to normalize activations between layers. This causes \textit{`spot'} like artifacts when trained to progressively increase the temporal compression (Figure \ref{fig:custom_gn}(a)). This occurs because the model tends to encode important global information in these high-norm latent pixels, as also observed in~\cite{polyak2024movie}. 
To address this phenomenon, we modify the group-normalization operation by removing the mean subtraction following \cite{karras2020analyzing, karras2022elucidating}. We found this effectively resolves the \textit{`spot'} artifacts in the reconstructed videos (Figure \ref{fig:custom_gn}(b)). 


\subsection{Progressive model growing}
\label{sec:growing}

\noindent \textbf{Motivation.}\label{sec:motivation} To increase the temporal compression of \lio from \fourx to \eightx or \sixteenx, the standard approach is to add additional downsampling and upsampling blocks in the bottleneck layers of the encoder and decoder respectively. In our preliminary study, we found that directly training \magvit for \sixteenx temporal compression leads to poor reconstruction quality for a $24$-fps video (Figure \ref{fig:interpolation}(b)). However, we observed that the \fourx temporal compression model can still accurately reconstruct a $6$-fps video by feeding the same $24$-fps video after subsampling frames by a factor of $4\times$ (Figure \ref{fig:interpolation}(a)). 
Since our base model \lio at \fourx temporal compression is similar to \magvit, this conclusion is also applicable for \lio.

Since the \lio-\fourx can accurately reconstruct a $6$-fps video, our strategy is to use it as guidance for the additional bottleneck downsampling and upsampling blocks to achieve \eightx and \sixteenx temporal compression. This can be thought of as a guided video interpolation problem. 
Instead of forcing the bottleneck compression layers to learn the entire information flow, we aim to induce them to reuse the information of the \fourx blocks, and only learn the essential information needed to synthesize the in-between frames, which is necessary to reconstruct the full video.


\noindent \textbf{Progressive model growing.}\label{sec:pro_grow} Building on the above observation, we develop our progressive model growing framework for boosting the temporal compression in \lio. 
We describe in detail below our framework for growing our model from \fourx to \eightx temporal compression. Growing the model from \eightx to \sixteenx follows the same procedure.
\\
\noindent \textbf{Key-frame embedding.} A naive way to achieve \eightx temporal compression using the \fourx model, $\mathbf{E}_{4\times}$, is to subsample the input video frames $\mathbf{v}$ by a factor of two, $\mathbf{v}_{//2}$, and then encode the frames with $\mathbf{E}_{4\times}$. This can be referred to as learning the key-frame encodings $z_{\text{key}}$ of the subsampled keyframes.

\vspace{-1mm}
\begin{equation}
    \begin{aligned}
    &z_{\text{key}} = \mathbf{\hat{E}}_{4\times}(\mathbf{v}_{\text{//2}})
    \end{aligned}\label{eq:loss}
\end{equation}
\noindent where $\mathbf{\hat{E}}_{4\times}$ is the encoder blocks without the bottleneck $1\times1\times1$ layers in \reffig{method}.
\\
\noindent \textbf{Residual embedding.} Now, with the encodings to accurately reconstruct keyframes in the video, we only need to learn the information for the remaining frames. 
We denote $\overset{*}{z}$ as the embedding of entire video $\mathbf{v}$ upto the \fourx temporal compression blocks, $\mathbf{\hat{E}}_{4\times}$. To reach \eightx compression, we compress $\overset{*}{z}$ through the additional 2$\times$ downsampling block. However, we need to ensure that the downsampling blocks $\mathbf{\hat{E}}_{4\times\text{-}8\times}$ are aware of the information already present in $z_\text{key}$, as it only needs to preserve the remaining information to avoid redundancy. For this, we use adaptive group normalization (\texttt{AdaNorm}) to condition the intermediate latent, $\overset{*}{z}$, on the key-frame embeddings $z_\text{key}$ before passing it to $\mathbf{\hat{E}}_{4\times\text{-}8\times}$.

\begin{equation}
    \begin{aligned}
    &\overset{*}{z}_{\text{inter}} = \text{\texttt{AdaNorm}}(z_{\text{key}}, \mathbf{\hat{E}}_{4\times}(\mathbf{v}))\\
    &z_{\text{inter}} = \mathbf{\hat{E}}_{4\times\text{-}8\times}(\overset{*}{z}_{\text{inter}})
    \end{aligned}\label{eq:loss}
\end{equation}

\noindent We obtain the final latent $z$ as the linear combination of both $z_{\text{key}}$ and $z_{\text{inter}}$.

\begin{equation}
    \begin{aligned}
    &z = \texttt{Conv}_{1\times1\times1}(z_{\text{key}} + 
    z_{\text{inter}})
    \end{aligned}\label{eq:loss}
\end{equation}

\noindent For the decoder we only add a bottleneck upsampling block.








\noindent \textbf{Training strategy.} We obtain the base \fourx temporal compression model by training with (1) the video-level GAN loss $\mathcal{L}_{\text{GAN}}$~\cite{goodfellow2014generative}, (2) the $L_1$ reconstruction loss $\mathcal{L}_{\text{rec}}$, and (3) the KL-divergence loss $\mathcal{L}_{\text{KL}}$. 
\begin{equation}
    \begin{aligned}
    &\mathcal{L}_{4\times} = \mathcal{L}_{\text{rec}} + \lambda_{\text{KL}} \mathcal{L}_{\text{KL}} + \lambda_{\text{GAN}}\mathcal{L}_{\text{GAN}}
    \end{aligned}\label{eq:loss}
\end{equation}
\noindent We initialize the weights for the \eightx temporal compression model with the \fourx models and freeze the encoder and decoder blocks initialized from the \fourx model. At this stage, we only train the newly added blocks and the $1\times1\times1$ bottleneck layer using (1) the $L_1$ reconstruction loss $\mathcal{L}_{\text{rec}}$, and (2) the KL-divergence loss $\mathcal{L}_{\text{KL}}$. 
\noindent While GAN loss can slightly improve the FVD score in the reconstructed videos, it tends to introduce visual artifacts in challenging scenarios. Therefore, we do not use it at this stage, which also makes training with progressive growing significantly faster. Details about training efficiency are provided in the supplementary.

\subsection{High resolution video reconstruction}
\label{sec:tile}

Directly using our model, \lio, or \magvit to encode and decode videos at high resolution (\eg 540$\times$960), results in out-of-memory (OOM) errors due to the VRAM-intensive 3D convolution layers. 
We found the OOM issue arises only during decoding (i.e., high-resolution videos can be encoded as-is). With this in mind, we opted to encode the entire video at once and only spatially tile the encoded latent, decoding each tile separately. However, similar to tiling in RGB space, this approach results in artifacts, as seen in Figure \ref{fig:tiling} (a). In the latent tiling case, we found the issue to be a misalignment between the full, low-resolution video encoding and decoding regimen learned during training and the tiled decoding seen at inference. Specifically, we found that there was a meaningful difference in normalization statistics encountered during training and inference. 
To mitigate this discrepancy, we introduce a \textbf{layer-wise spatial tiling} technique: during decoding, we divide the input tensor to \emph{each} \texttt{conv} layer into overlapping spatial tiles, process each tile independently, and merge the tiles back together using linear interpolation weights to blend the overlapping components. Using this approach, we can decode into high-resolution videos without any artifacts (Figure \ref{fig:tiling} (b)). 

\input{figures/tiling}

%% file: figures/method.tex
\begin{figure*}[t!]\vspace{-4mm}
    \centering
    \includegraphics[width=0.9\linewidth]{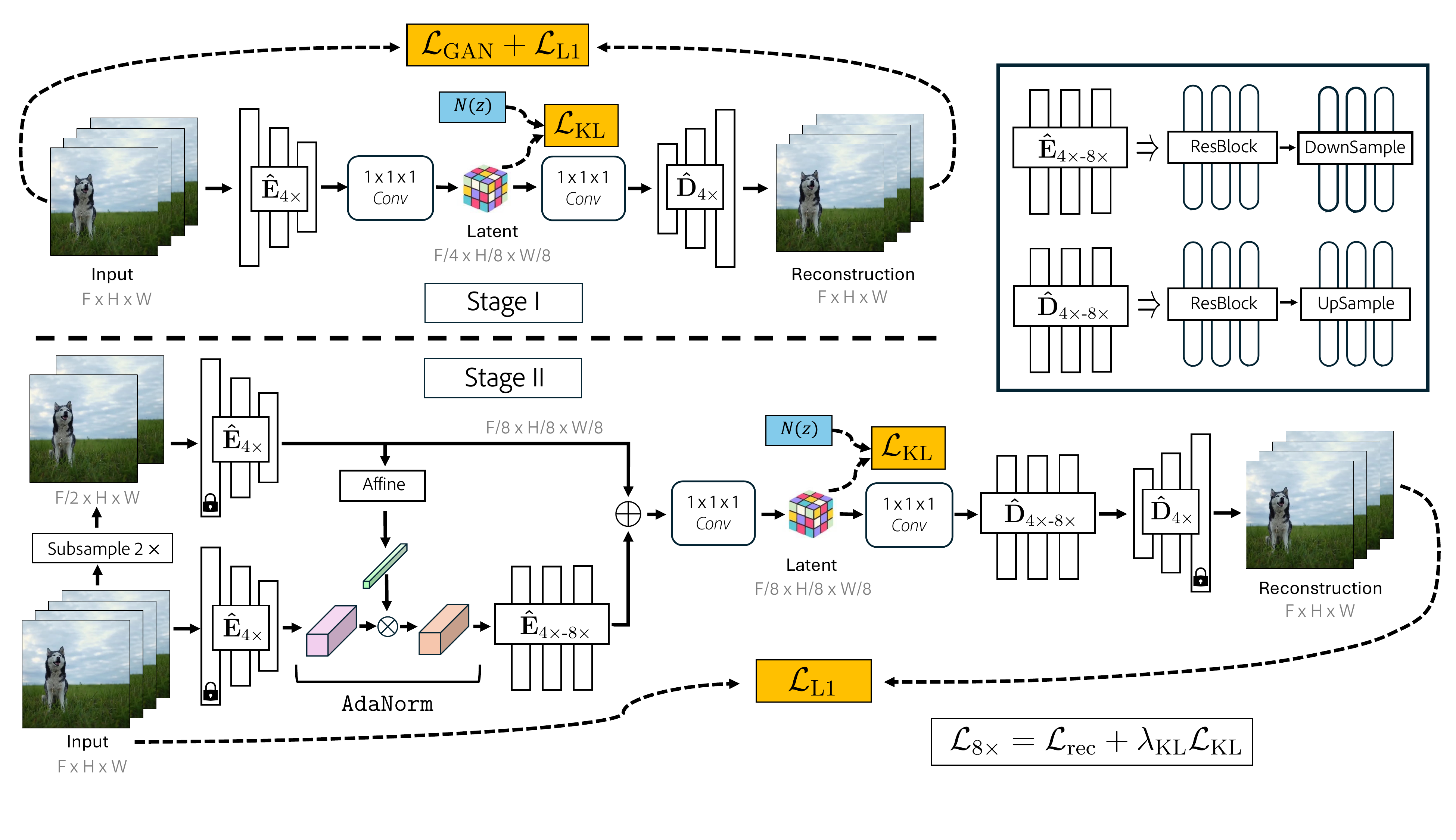}
    \vspace{-7mm}
    \caption{\camera{{\bf Methodology.} Figure shows details of our method of progressive growing. In Stage I (top) we show a method of training our base \fourx video tokenizer. (bottom) we illustrate the detailed method of growing the base \fourx model to achieve \eightx temporal compression.}}
    \lblfig{method}
\vspace{-4mm}
\end{figure*}

%% file: figures/tiling.tex
\begin{figure}[t!]
    \centering
    \includegraphics[width=\linewidth]{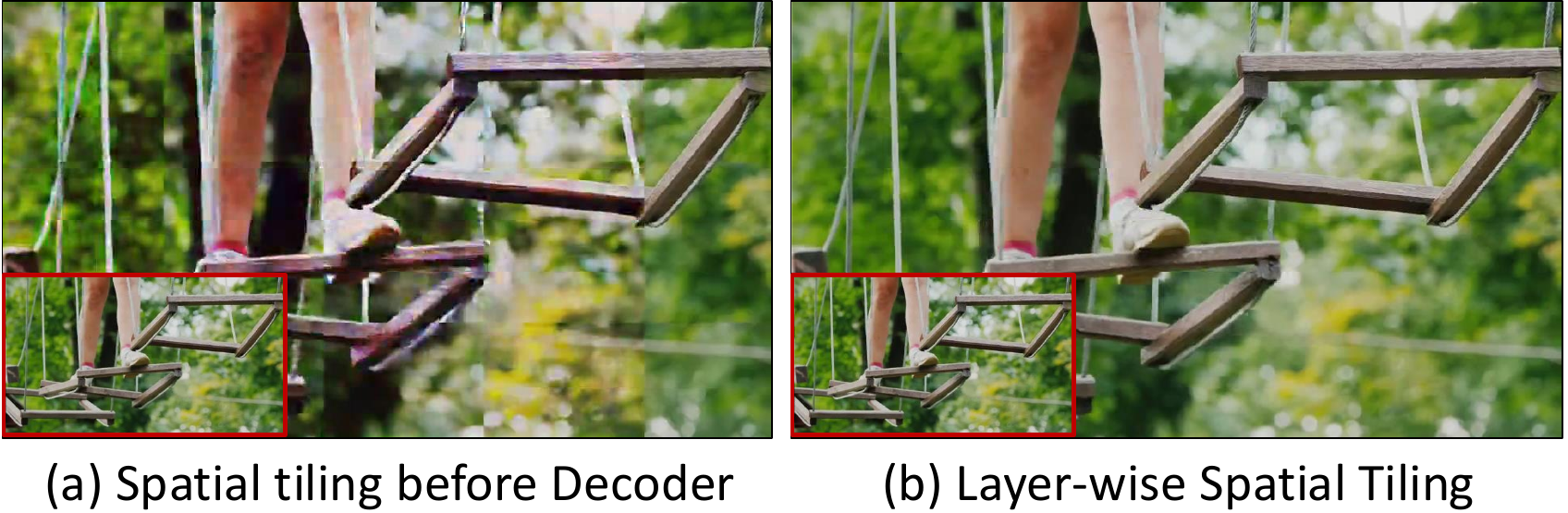}
    \vspace{-4mm}
    \caption{\camera{{\bf High-Resolution Reconstruction.} (left) tilting the latent before passing through the decoder leads to artifacts in the reconstructed video. (right) layer-wise tiling resolves the artifacts. The \textcolor{red}{red-bordered} frame at the bottom left corner represents the ground-truth frame.}}
    \lblfig{tiling}
\vspace{-4mm}
\end{figure}

%% file: Experiments.tex
\subsection{Reconstruction Quality}

\input{figures/reconstruction_baseline}

\noindent \textbf{Implementation details.} We train our models on our internal dataset of 300M images and 15M videos at 256$\times$256 resolution. The \fourx, \eightx, and \sixteenx temporal compression models are trained with 17-frame training video clips sampled at 6 fps, 12 fps, and 24 fps, respectively. Since we follow \magvit's approach of using 3D \textit{causal convolution} layers for spatiotemporal processing, videos with $1 + k \text{ x } N$ frames are compressed into $1 + N$ latent frames, where $k$ is the compression ratio. Following this, the \fourx models compress videos from 17 frames to 5 latent frames, similarly, \eightx from 17 frames to 3, and, \sixteenx from 17 frames to 2.

\noindent \textbf{Baselines.} We compare our method to the following state-of-the-art methods:
\begin{itemize}
    \item \magvit~\cite{yu2023language}. As the official codebase is unavailable, we implement the continuous space version ourselves and train with a setting similar to our method. This is our main baseline where we perform full-model training for \fourx, \eightx, and \sixteenx temporal compression at $z\_dim=$ 8 and 16.
    \item For fairness of comparison, we compare with SOTA tokenizers with $z\_dim=$ 8 and 16. We compare against OmniTokenizer~\cite{wang2025omnitokenizer}, VidTok~\cite{tang2024vidtok}, Cosmos-CV (continuous version)~\cite{agarwal2025cosmos}, VAE in CogVideoX~\cite{yang2024cogvideox}, Wan~\cite{wan2025}, Hunyuan Video~\cite{huang2018multimodal}, CV-VAE~\cite{zhao2025cv}, and, WF-VAE~\cite{li2024wf}. All the tokenizers are continuous-space, except for OmniTokenizer, which is discrete. Additionally, all the open-sourced SOTA tokenizers operate at \fourx temporal compression, except Cosmos-CV, which has both \fourx and \eightx temporal compression.
\end{itemize}

\input{tables/reconstruction_4x}

\noindent We compare the above baselines against our base \lio-\fourx temporal compression. For more aggressive compression, \eightx and \sixteenx, we extend and train \magvit at the target compression ratio and compare against our method.

\noindent \textbf{Evaluation benchmarks.} We evaluate all methods for reconstruction quality on two standard video benchmarks, MCL-JCV~\cite{wang2016mcl} and DAVIS 2019  (Full Resolution)~\cite{caelles20192019}. 
We select two different settings for reconstruction, one at square crops of 512$\times$512 and the other at 360$\times$640 (at native aspect ratio of 16:9). We compare the reconstruction quality on PSNR, LPIPS~\cite{zhang2018perceptual}, and Fr\'echet video distance (FVD)~\cite{unterthiner2018towards, ge2024content}.


\input{tables/reconstruction_high}

\noindent \textbf{Effectiveness of our base \fourx \lio.} While our main goal is not to achieve the best \fourx temporal compression model, we do this experiment to verify that the base \fourx model we developed can serve as a strong tokenizer.
From Table \ref{tab:reconstruction_4x}, we see that \magvit and \lio at $z\_dim=8$ perform better than Omnitokenizer~\cite{wang2025omnitokenizer} and are comparable to SOTA tokenizer VidTok~\cite{tang2024vidtok}. Thus, we build our base model inspired from \magvit architecture. Even at $z\_dim=16$ our method performs better than SOTA methods like Cosmos-CV (continuous)~\cite{agarwal2025cosmos}, CogVideoX~\cite{yang2024cogvideox}, and CV-VAE~\cite{zhao2025cv} across all metrics and datasets and comparable to very recent SOTA methods like VidTok~\cite{tang2024vidtok} WF-VAE~\cite{li2024wf}, Wan~\cite{wan2025}, and Hunyuan Video~\cite{kong2024hunyuanvideo}, which focus on creating high-quality \fourx temporal compression video tokenizers. Thus, our base \fourx tokenizer is a competitive model in terms of reconstruction quality. 

\noindent \textbf{Effectiveness of progressive growing for boosting temporal compression.} We compare our method, \lio, against \magvit on 2 different settings, one with 8 channels and the other with 16 channels in the latent dimension for both \eightx and \sixteenx temporal compression. From Table~\ref{tab:reconstruction_high}, \lio achieves better reconstruction quality than \magvit directly extended to high compression. It is interesting to note that the 16-channel version of \lio-\eightx can maintain a somewhat similar reconstruction quality to \lio-\fourx. Qualitative analysis in \reffig{baseline_rec}, (left) we observe that at \sixteenx compression, \magvit is unable to preserve even the coarse structural details in the face or the dog or the human hand. Even at \eightx temporal compression, in regions of very high motion like the cyclist, \magvit blends the foreground object with the background, causing noticeable motion blur in the reconstructed videos. \lio at \eightx temporal compression also performs better than Cosmos-CV.

\noindent \textbf{Ablation on using GAN loss for progressive growing.} In this ablation study, we verify whether adding GAN loss during progressive growing is beneficial. From Table~\ref{tab:reconstruction_gan}, we find GAN loss at this stage while slightly improving perceptual quality (LPIPS and FVD), also slightly reduces PSNR. Additionally, not using GANs while progressive growing improves training efficiency (details in supplementary). Thus, we choose not to GAN loss during progressive growing.













\input{figures/t2v_eval}

\subsection{T2V Quality with Compressed Latent Space}
\label{t2v_exp}

\input{tables/reconstruction_gan}

Our ultimate goal is to generate a latent space that is not only compact but more importantly, good for downstream diffusion model training.
We evaluate the latent spaces of our tokenizers by accessing the text-to-video generation quality of the Diffusion Transformer (DiT) model trained with the resulting latent spaces. 
In this section, we evaluate our \fourx and \sixteenx models. We aim to investigate if DiT training quality and convergence degrades when trained with the highly compressed latent space of \sixteenx temporal compression compared to the commonly used \fourx temporal compression. 

\noindent \textbf{Training and implementation details.} Our text-to-video diffusion model is based on the standard DiT formulation~\cite{peebles2023scalable}, composed of multiple Transformer blocks, where we replace spatial self-attention with spatial-temporal self-attention blocks.
We train our model on an internal dataset of images and videos. We first train DiT on 256p images for 200K iterations and then jointly train on images and videos for about 150K more iterations.

\input{figures/t2v_example}

\noindent \textbf{Evaluation metrics.} We evaluate the quality of text-to-video generation using VBench~\cite{huang2024vbench}. Following the official guidelines, we generate videos with all the 946 provided text prompts and generate videos with 5 random seeds per prompt. We evaluate based on all 16 quality dimensions.

\noindent \textbf{T2V generation quality.} From \reffig{t2v_efficiency}(b), we find that training DiT with \sixteenx temporally compressed latent space does not degrade the video generation quality on almost all dimensions compared to \fourx compression latent. 
\reffig{t2v_results} shows two diverse and vibrant videos generated by DiT with \sixteenx compressed latent. We find that the DiT can accurately follow the caption and generate very realistic motion.

\noindent \textbf{Efficiency.} 
Using a compact latent space allows DiT to operate with fewer tokens, significantly reducing GPU memory and time. In a fixed frame length case (68 frames, $\sim$2.8s at 24 fps), DiT with \fourx compression generates 20 latents at 0.99s/timestep, while \eightx compression reduces this to 12 latents at 0.58s (1.7$\times$ speedup). With \sixteenx, only 8 latents are needed, achieving 0.39s/timestep (2.5$\times$). In a fixed token budget of 40 latent frames, the \fourx model generates a 136-frame video (~5.6s), whereas the \sixteenx model produces 340 frames (~14.1s), a 2.5$\times$ increase in video length.







%% file: figures/reconstruction_baseline.tex
\begin{figure*}[t!]\vspace{-5.5mm}
    \centering
    \includegraphics[width=0.98\linewidth]{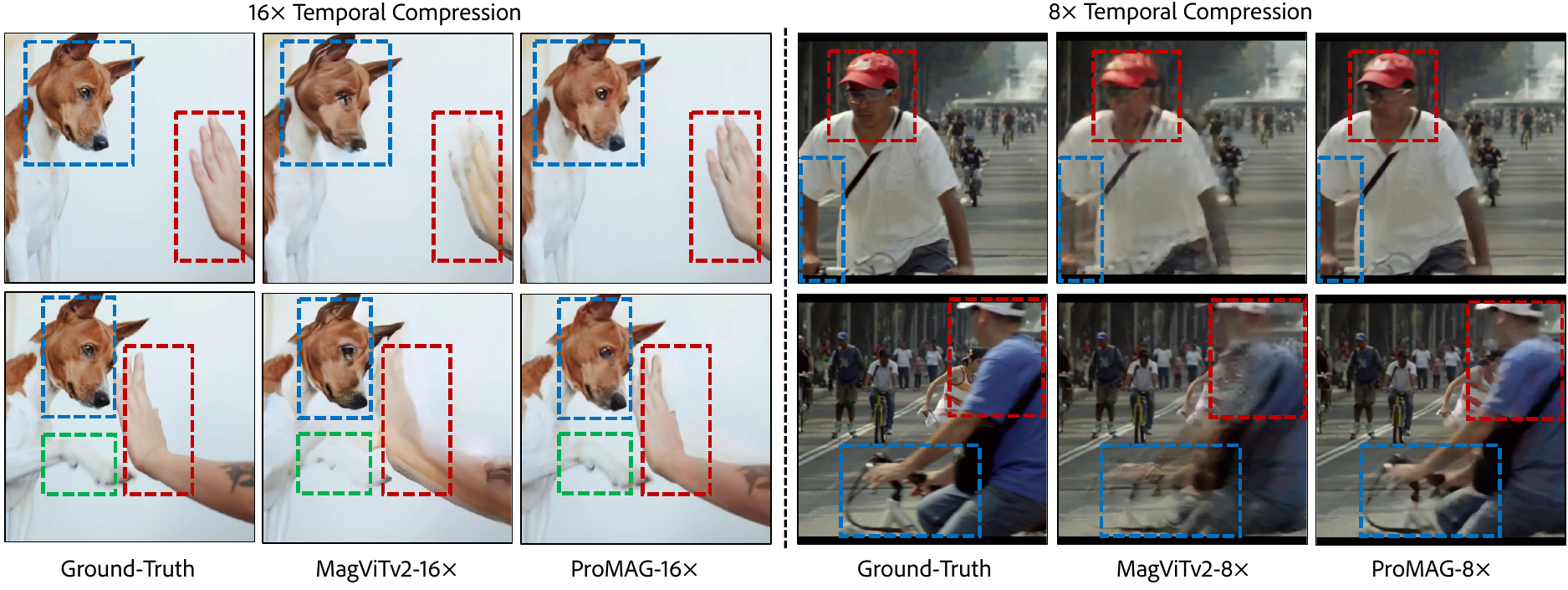}
    \vspace{-3mm}
    \caption{\textbf {Effectiveness of Progressive Growing at high temporal compression.} Video reconstruction example for (left) \sixteenx temporal compression, (right) \eightx compression. \magvit causes loss of details and produces artifacts in the reconstructed frames, like motion blur in the dog's and human's hand (left) and cyclist (right). The dotted boxes show regions of the most difference.}
    \lblfig{baseline_rec}
\vspace{-2mm}
\end{figure*}

%% file: tables/reconstruction_4x.tex
\begin{table}[!t]
\vspace{-2mm}
\centering
\setlength{\tabcolsep}{3pt}  
\scalebox{0.7}{
\begin{tabular}{lccccccccc}
\toprule
\multirow{2}{*}{Method} & \multirow{2}{*}{$z\_dim$} &
\multicolumn{3}{c}{MCL-JCV} &
\multicolumn{3}{c}{DAVIS} \\
\cmidrule(lr){3-5}\cmidrule(lr){6-8}
& & PSNR & LPIPS & rFVD & PSNR & LPIPS & rFVD \\
\midrule
MagViT-v2 & \multirow{4}{*}{8} &
30.63 & 6.49 & 58.11 & 28.32 & 7.30 & 129.11 \\
OmniTokenizer~\cite{wang2025omnitokenizer} & &
24.55 & 11.06 & 124.83 & 23.60 & 15.02 & 290.38 \\
VidTok~\cite{tang2024vidtok} & &
\textbf{31.20} & 6.93 & 62.41 & \textbf{28.73} & 8.05 & 150.54 \\
\lio & &
30.99 & \textbf{6.53} & \textbf{57.91} & 28.43 & \textbf{7.38} & \textbf{130.56} \\
\midrule
Cosmos-CV~\cite{agarwal2025cosmos} & \multirow{6}*{16} & 32.08 & 9.54 & 50.65 & 29.81 & 13.71 & 113.13\\
CogVideoX~\cite{yang2024cogvideox}  &  & 32.57 & 6.21 & 50.75 & 29.8 & 7.49 & 100.56\\
VidTok~\cite{tang2024vidtok} &  & 32.95 & 5.55 & 37.7 & \textbf{31.12} & 5.22 & 94.04\\
WF-VAE~\cite{li2024wf} &  & \textbf{33.18} & 5.09 & 39.71 & 30.29 & 6.51 & 88.27\\
Wan~\cite{wan2025} & & 32.27 & 5.15 & 39.33 & 30.15 & 5.29 & 98.49\\
Hunyuan~\cite{kong2024hunyuanvideo} & & 33.13 & 4.95 & 36.21 & 30.4 & 5.02 & 97.56\\
CV-VAE~\cite{zhao2025cv} &  & 32.83 & 6.31 & 46.35 & 30.07 & 7.93 & 101.94 & \\
\lio &  &  32.94 & \textbf{4.91} & \textbf{34.21} & 30.77 & \textbf{4.98} & \textbf{93.92}\\
\bottomrule
\end{tabular}
}
\caption{\textbf{Reconstruction comparison at base \fourx temporal compression.}
We compare our method with baselines on MCL-JCV and DAVIS under 8$\times$8$\times$4 compression on 512$\times$512 resolution.}
\vspace{-4mm}
\label{tab:reconstruction_4x}
\end{table}

%% file: tables/reconstruction_high.tex
\begin{table}[!t]
\vspace{-2mm}
\centering
\setlength{\tabcolsep}{3pt}  
\scalebox{0.67}{
\begin{tabular}{lccc ccc ccc}
\toprule
\multirow{2}{*}{Method} & \multirow{2}{*}{Compression} & \multirow{2}{*}{$z\_dim$} & \multicolumn{3}{c}{MCL-JCV} & \multicolumn{3}{c}{DAVIS} \\
\cmidrule(lr){4-6}\cmidrule(lr){7-9}
& & & PSNR & LPIPS & rFVD & PSNR & LPIPS & rFVD \\
\midrule

\magvit & \multirow{2}{*}{8$\times$8$\times$8} & \multirow{2}{*}{8} & 28.47 & 11.17 & 219.36 & 24.02 & 16.7 & 525.1\\
\lio    &                                       &                      & \textbf{30.26} & \textbf{8.36} & \textbf{96.85}  & \textbf{26.48} & \textbf{11.37} & \textbf{360.54}\\
\midrule
\magvit & \multirow{2}{*}{8$\times$8$\times$16} & \multirow{2}{*}{8} & 26.15 & 14.54 & 255.51 & 21.82 & 20.54 & 728.04\\
\lio    &                                        &                      & \textbf{28.31} & \textbf{11.2} & \textbf{183.4} & \textbf{23.89} & \textbf{16.9} & \textbf{608.43}\\
\midrule
Cosmos-CV~\cite{agarwal2025cosmos} & \multirow{3}{*}{8$\times$8$\times$8} & \multirow{3}{*}{16} & 31.05 & 11.79 & 101.36 & 28.27 & 17.87 & 222.61\\
\magvit                             &                                   &                      & 28.51 & 10.62 & 135.57 & 27.18 & 8.84  & 243.19\\
\lio                                &                                   &                      & \textbf{32.06} & \textbf{6.94}  & \textbf{63.64}  & \textbf{28.70} & \textbf{8.32}  & \textbf{194.79}\\
\midrule
\magvit                             & \multirow{2}{*}{8$\times$8$\times$16} & \multirow{2}{*}{16} & 28.51 & 10.62 & 135.57 & 24.17 & 14.84 & 427.52\\
\lio                                &                                   &                      & \textbf{30.02} & \textbf{9.24}  & \textbf{115.25} & \textbf{25.64} & \textbf{13.89} & \textbf{320.80}\\
\bottomrule
\end{tabular}}
\caption{\textbf{Reconstruction comparison at high temporal compression.}
We compare our method with baselines on MCL-JCV and DAVIS under 8$\times$8$\times$4 compression on 512$\times$512 resolution.}
\label{tab:reconstruction_high}
\vspace{-4mm}
\end{table}

%% file: figures/t2v_eval.tex
\begin{figure}[t!]
    \centering
    \includegraphics[width=\linewidth]{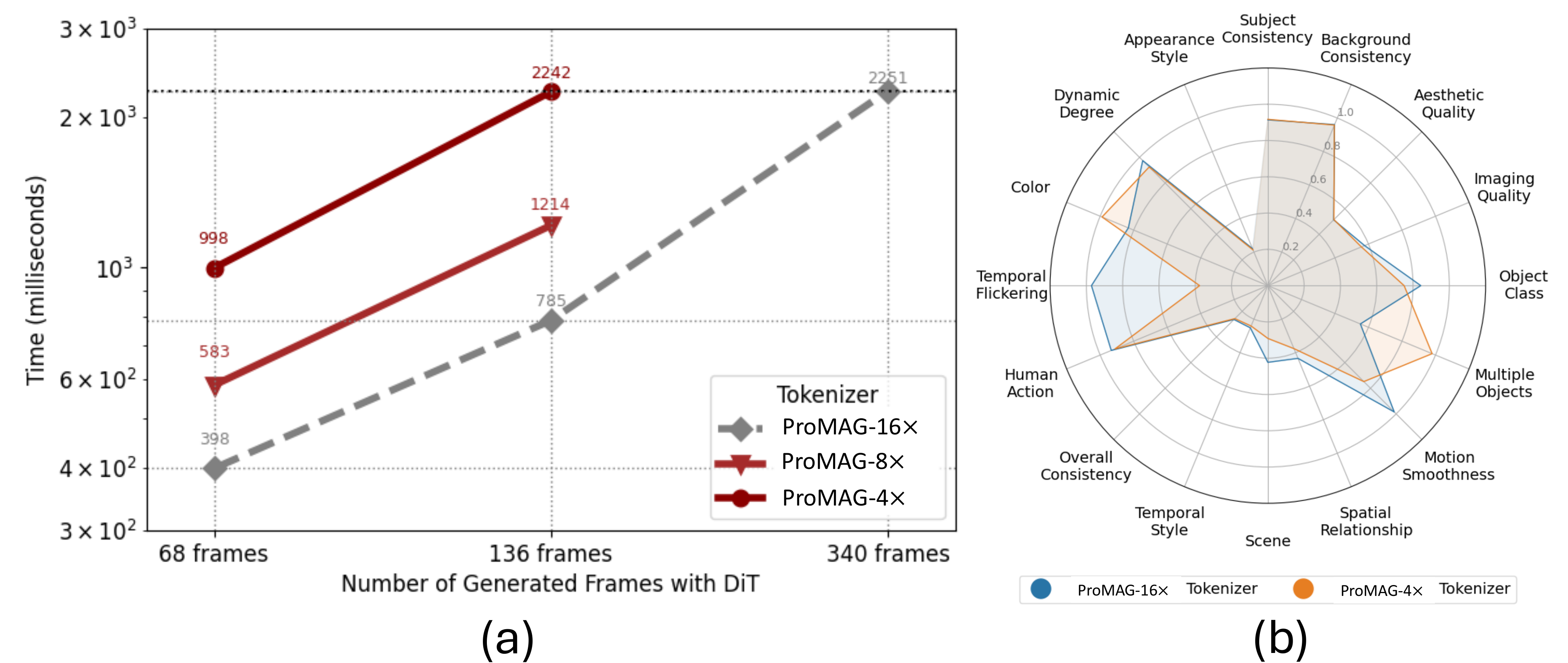}
    \vspace{-5mm}
    \caption{\camera{{\bf Text-to-Video Evaluation.} Quantitative comparison of T2V generation model train on \fourx and \sixteenx latent space of \lio. (top) The graph of time taken per denoising step of the diffusion model for generating N frame video. The time is computed on H100 with Flash Attention 3~\cite{shah2024flashattention}. The resolution of the videos is $192\times360$ pixels. We find that it takes the same time per diffusion step to generate a 136-frame video using the \fourx latent space, as generating a 340-frame video using the \sixteenx latent space. (bottom) Quantitative analysis different dimension on VBench~\cite{huang2024vbench} on generated videos with \fourx and \sixteenx compressed latents.}}
    \lblfig{t2v_efficiency}
\vspace{-3mm}
\end{figure}

%% file: tables/reconstruction_gan.tex
\begin{table}[!t]
\centering
\setlength{\tabcolsep}{3pt}  
\scalebox{0.64}{
\begin{tabular}{lccc ccc ccc}
\toprule
\multirow{2}{*}{Method} & \multirow{2}{*}{Compression} & \multirow{2}{*}{$z\_dim$} & \multicolumn{3}{c}{MCL-JCV} & \multicolumn{3}{c}{DAVIS} \\
\cmidrule(lr){4-6}\cmidrule(lr){7-9}
& & & PSNR & LPIPS & rFVD & PSNR & LPIPS & rFVD \\
\midrule

\lio & \multirow{2}{*}{8$\times$8$\times$8} & \multirow{2}{*}{8} & \textbf{32.06} & 6.94  & 63.64  & \textbf{28.70} & 8.32  & 194.79\\
\lio (w/ GAN)                      &                                   &                      & 31.78 & \textbf{6.21} & \textbf{53.18} & 28.21 & \textbf{7.76} & \textbf{160.22}\\
\midrule
\lio                             & \multirow{2}{*}{8$\times$8$\times$16} & \multirow{2}{*}{16}          & \textbf{30.02} & 9.24  & 115.25 & \textbf{25.64} & 13.89 & 320.80\\
\lio (w/ GAN)                      &                                   &                      & 29.67 & \textbf{8.38} & \textbf{99.64}  & 25.05 & \textbf{12.04} & \textbf{274.24}\\
\bottomrule
\end{tabular}}
\caption{\textbf{Ablation on GAN loss during progressive growing.}
We compare our method of progressive growing with and without using GAN loss.}
\label{tab:reconstruction_gan}
\vspace{-4mm}
\end{table}

%% file: figures/t2v_example.tex
\begin{figure}[t!]
    \centering
    \includegraphics[width=\linewidth]{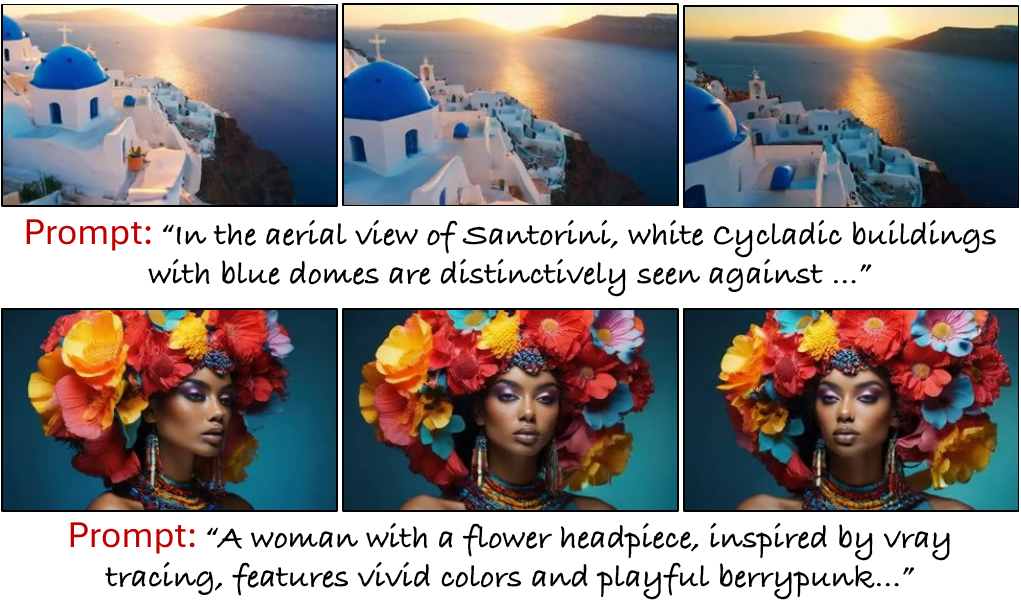}
    \vspace{-5mm}
    \caption{\camera{{\bf Text-to-Video Generation Results.} } We show vibrant videos generated by DiT on our \lio-\sixteenx latent space. This highlights that training DiT on highly compressed latent space can generate videos with accurate text coherence and realistic motion.}
    \lblfig{t2v_results}
\vspace{-4mm}
\end{figure}

%% file: Discussion.tex
\subsection{Progressive growing vs. progressive training}

\input{tables/reconstruction_ablation}

\noindent We want to investigate if the effectiveness of our method comes from our specific design towards reusing pretrained high-quality, lower compression model and only learning the remaining information in the bottleneck layers, or simply just from progressive training of the model in stages by adding more bottleneck layers (w/o skip information flow). 

\noindent \textbf{Progressive training.} In this case, we start from a pretrained \fourx model, and then add bottleneck downsampling and upsampling blocks to the encoder and decoder respectively, to grow the compression rate and only train the additional blocks. This does not include the use of subsampled frames encoding or \texttt{AdaNorm} as our full method (Figure \ref{fig:method}). Progressive training, in essence, is similar to tokenizer in OpenSORA~\cite{opensora} and OpenSORA-Plan~\cite{pku_yuan_lab_and_tuzhan_ai_etc_2024_10948109}, which first trains a spatial-only VAE and then adds additional bottleneck \fourx temporal compression blocks.

\noindent \textbf{Evaluation.} Table \ref{tab:ablation} shows the comparison of the reconstruction quality on MCL-MJC at 512x512 crops for \eightx temporal compression. We find that progressive training (Table \ref{tab:ablation}, $2^\text{nd}$ row) by itself, improves the PSNR, LPIPS, and FVD compared to directly training \magvit from scratch for \eightx compression. We hypothesize that it might be because with large compression and increasing channel multiples, training the video tokenizer at once may lead to difficulty in convergence. On top of that, our method of progressive growing further improves the reconstruction of all the metrics.




\subsection{Progressive growing vs. frame subsampled encoding + external interpolation}

\input{figures/external_interpolation}
\input{tables/interpolation}

The primary motivation for our method of progressive growing was that the \fourx compression model performs very accurate reconstruction even for a 6fps video (after 4 times frame subsampling from a 24fps video).

\noindent \textbf{Baseline.} A very natural question would be how our method for \eightx (or \sixteenx) temporal compression would fare against the solution where we first perform \fourx temporal compression on videos subsampled with a factor of 2 (or 4), followed by using external state-of-the-art frame interpolation method like EMA-VFI~\cite{zhang2023extracting} to perform 2X (4X) frame interpolation on the reconstructed frames. We observe, in \reffig{external_interpolation} that, performing external interpolation in regions of very high and complex motion causes much more blurring compared to applying our \lio-\sixteenx directly on the original videos.




\noindent \textbf{Evaluation.} From Table \ref{tab:interpolation}, we see that our method for \eightx temporal compression achieves much better reconstruction quality compared to using our \fourx model on subsampled frames followed by 2X interpolation. A similar trend can be observed in the case of \sixteenx temporal compression.



%% file: tables/reconstruction_ablation.tex
\begin{table}[t]
\centering
\scalebox{0.7}{
\begin{tabular}{lccc}
\toprule
\multirow{2}*{Method} & \multicolumn{3}{c}{MCL-JCV}\\
\cmidrule(lr){2-4}
& PSNR & LPIPS & rFVD\\
\midrule
\magvit & 28.51 & 10.62 & 135.57\\
\lio (w/o residuals \& \texttt{AdaNorm}) & 30.35 & 8.05 & 88.18 \\
\lio (full method) & \textbf{32.06} & \textbf{6.94} & \textbf{63.64}\\

\bottomrule
\end{tabular}}
\caption{\textbf{Progressive Growing v/s Progressive training}. Comparison of reconstruction quality of our full method against the naive way of progressive training and \lio on the 512x512 crops of the MCL-JCV video dataset.
}
\label{tab:ablation}
\end{table}

%% file: figures/external_interpolation.tex
\begin{figure}[t!]
    \centering
    \includegraphics[width=0.9\linewidth]{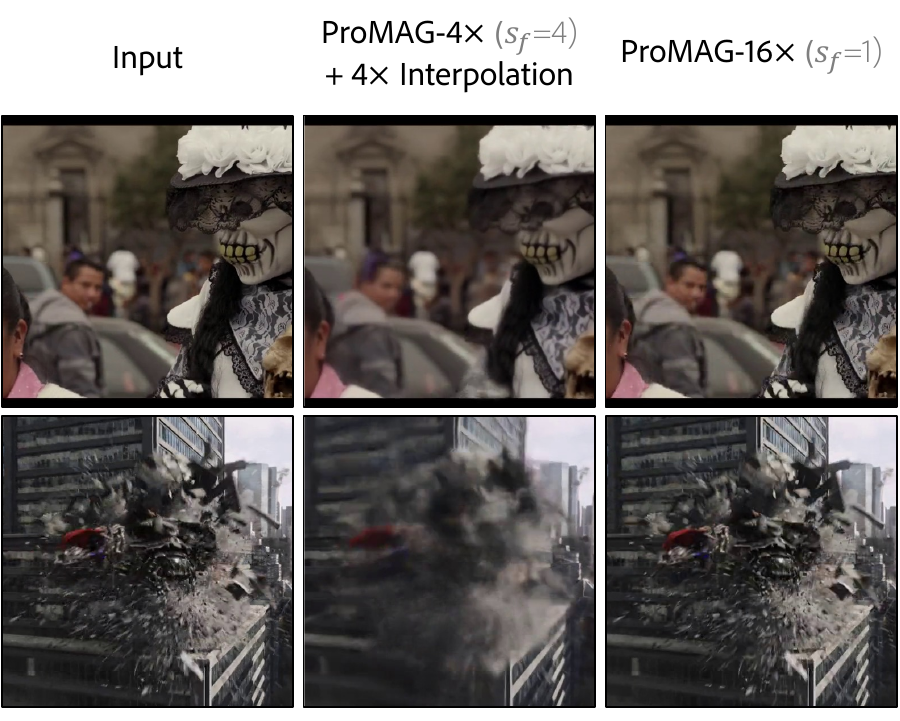}
    \vspace{-2mm}
    \caption{\textbf{Qualitative Comparison with External Interpolation.} Reconstruction comparison of our method \lio with \sixteenx temporal compression on a 24fps video, against a baseline where we first encode the video at low fps (with frame subsampling $s_f=4$), in this case 6fps, followed by using external interpolation method to generate the 4 in-between frames. (middle) The baseline with external frame interpolation produces very blurry outputs in regions of abrupt and high-intensity motion, compared to our method \lio-\sixteenx operating on 24fps video directly (right). This implies in some sense, that our progressive growing approach has in some sense learned a better interpolation for reconstruction.}
    \lblfig{external_interpolation}
\vspace{-4mm}
\end{figure}

%% file: tables/interpolation.tex
\begin{table}[t]
\centering
\scalebox{0.7}{
\begin{tabular}{lccc}
\toprule
\multirow{2}*{Method} & \multicolumn{3}{c}{MCL-JCV}\\
\cmidrule(lr){2-4}
& PSNR & LPIPS & rFVD\\
\midrule
\lio (\fourx w/ $s_f=2$ + 2X Interpolation) & 30.68 & 8.61 & 230.16 \\
\lio (\eightx w/ $s_f=1$) & \textbf{32.06} & \textbf{6.5} & \textbf{53.77}\\
\hdashline
\lio (\fourx w/ $s_f=4$ + \fourx Interpolation) & 28.42 & 12.35 & 341.14 \\
\lio (\sixteenx w/ $s_f=1$) & \textbf{29.6} & \textbf{9.93} & \textbf{147.48}\\

\bottomrule
\end{tabular}}
\caption{\textbf{Quantitaive Comparison with External Interpolation}. Reconstruction quality on MCL-JCV dataset on 512x512 crops on \eightx and the other at \sixteenx temporal compression.
}
\label{tab:interpolation}\vspace{-4mm}
\end{table}

%% file: Conclusion.tex
In this work, we push the boundary to which we can perform temporal compression while keeping latent channel dimension constant. To this end, we propose a novel method of progressively growing our \fourx temporal compression video tokenizer, \lio, to \eightx and subsequently \sixteenx temporal compression. Through extensive evaluation of reconstruction and text-to-video generation, we show that (i) our video tokenizer can perform much better reconstruction than baseline methods at very high temporal compression rates, and (ii) our high compression latent space is suitable for DiT training and provides an immense boost in terms of efficiency for video generation. However, there still exist limitations, which offer opportunities for future research. In scenarios of very high and abrupt motion, like sports videos, our high-compression tokenizers, while performing significantly better than the directly trained baselines, also suffer from temporal artifacts in reconstruction. 

%% file: appendix.tex
\section{Progressive Growing - from \eightx to \sixteenx Temporal Compression}

\input{figures/method_supp}
\noindent In Figure 2 of the main paper we illustrate our method of growing the temporal compression of the \fourx video tokenizer to \eightx temporal compression. In \reffig{method_supp}, we show the details of our entire method, i.e., growing the base \fourx model to \eightx then ultimately to \sixteenx temporal compression in 3 stages. Note that all the parameter weights in the subsequent are initialized from their corresponding parameters in the previous stage. The newly added parameters are initialized with random weights.

\section{\lio Implementation Details}

\noindent \textbf{Model Architecture. }In Table \ref{tab:hyperparameters}, we describe the details of our tokenizer architecture for \fourx, \eightx and \sixteenx temporal compression models. The encoder and decoder design of our model, \lio follows \magvit structure. The discriminator is taken from Stable Diffusion VAE~\cite{rombach2022high}, where we replace the \texttt{Conv2D} with \texttt{Conv3D}. In our case, using the \magvit discriminator produced checkerboard-like artifacts in reconstunction.

\noindent \textbf{Training Hyperparameters. }We provide additional detailed training hyper-parameters for our models as listed below:

\begin{itemize}
    \item Video input: 17 frames, frame stride 1, $256 \times 256$ resolution.
    \item Reconstruction loss weight: 1.0.
    \item Generator loss type: Hinge Loss~\cite{lim2017geometric}.
    \item Generator adversarial loss weight: 0.1 for \fourx and 0.0 for \eightx and \sixteenx (Note: we do not use discriminator to progressively grow the model to \eightx and \sixteenx temporal compression.)
    \item Discriminator gradient penalty: 0.0 r1 weight.
    \item VGG Perceptual loss weight: 1.0.
    \item KL-Divergence loss weight: 1e-12.
    \item Tokenizer Learning rate: 0.0001.
    \item Tokenizer Optimizer Params: Adam with $\beta_1 = 0.9$ and $\beta_2 = 0.99$.
    \item Tokenizer weight decay: 0.0001
    \item Discriminator Learning rate: 0.0001.
    \item Discriminator Optimizer Params: Adam with $\beta_1 = 0.9$ and $\beta_2 = 0.99$.
    \item Discriminator weight decay: 0.0001
    \item Training iterations: 300K for \fourx model, 100K for \eightx model (on top of \fourx model), 100K for \sixteenx model (on top of \eightx model).
    \item Global Batch size: 32.
\end{itemize}

\section{Text-to-Video Implementation Details}

\noindent \textbf{Model Architecture.} Our text-to-video
diffusion model is based on the standard DiT~\cite{peebles2023scalable}, composed of multiple Transformer blocks, where
we replace spatial self-attention with spatial-temporal self-
attention blocks. The model architecture is similar to the diffusion transformer in CogVideoX~\cite{yang2024cogvideox}. Following them, we also keep the number of model parameters to be around 5B.

\noindent \textbf{Dataset Details.} We train on our internal dataset of 300M images and 1M videos. Images contain different aspect ratios, while videos are of 192x360 resolution. We train using a relatively small scale and low resolution since the computation cost for training text-to-video models is very enormous. Additionally, our main focus in training the text-to-video model is to verify that our extremely compressed latent space (\sixteenx temporal compression) is compatible with DiT training and can achieve a similar quality to \fourx compressed latent space. Our goal is not to compete with state-of-the-art text-to-video models.

\noindent \textbf{Training Details. } Following standard practices~\cite{singer2022make, ho2022imagen, ge2023preserve, gupta2025photorealistic}, we train our text-to-video diffusion model in 2 stages. In the first stage (image pertaining), we train the model only on images. In the second stage (joint training), we train the model jointly on both images and videos. We train the first stage for about 200K iterations and the second stage for around 150K additional iterations. We train our models on 8 nodes of H100 (64 GPUs in total).

\noindent \textbf{Training on Longer Videos. } In Section 4.2 (Efficiency) in the main paper, we discuss that due to the highly compact nature of our \sixteenx latent space, we can use the same token budget for 340 frames (= 20 latent frames for \sixteenx temporal compression), which is the same as 136 frames with \fourx temporal compression. To verify the generation quality of very long videos 340 frames (or 14.1s) video at 24fps, we also train text-to-video model for this. More specifically, we only finetune our text-to-video model trained on \sixteenx temporally compressed latent space for an additional 10K iterations on video training of 340 frames. 

\section{Resolving Jump Issues: Overlapping Frame Reconstruction and Text-to-Video Generation}
\noindent Following \magvit~\cite{yu2023language}, train \lio on 17 frame video, for all \fourx, \eightx and \sixteenx temporal compression ratio. We found that since we train the model on 17 frames, we can only do encoding and decoding with 17 (or less) frames. We found the similar case is true with \magvit. We found that reconstruction of more than 17 frames causes deterioration in the reconstruction results beyond 17 frames, like blurring, or in some cases checkerboard-like artifacts. Thus for reconstruction (or text-to-video generation), for an N frame video, we need to process it in chunks of 17 frames at a time. This causes jumps like effect in regions of high frequency details every 17 frames. This jumps is much less noticeable in reconstruction results, but more noticeable in the text-to-video generation results. For video generation this jumping effect reduces with more training iterations but still persists slightly. To counteract this, we perform encoding in chunks with overlap of few frames (in this case of 4 frames) \reffig{overlapping}. The overlapped regions in the output is blended in pixel-space with linear interpolated weights. This resolves the jumping artifacts perceptually both in reconstruction and video generation results (please refer to video results showing this in the supplementary videos).

\input{figures/overalapping}


\section{Training Efficiency of Our Approach}

\begin{table}[h]
\centering
\scalebox{0.7}{
\begin{tabular}{lcccc}
\toprule
\multirow{2}*{Method} & \multirow{2}*{Compression} & \multirow{2}*{Time/iter} & Total iters & Training time \\
& & & (cumulative) & (cumulative) \\
\midrule
\magvit & \multirow{2}*{$8\times8\times4$} & 0.42s & 300K & 1.45 days \\
\lio & & 0.42s & 300K & 1.45 days \\
\hdashline
\magvit & \multirow{2}*{8$\times$8$\times$8} & 0.92s & 400K & 4.25 days \\
\lio (4$\times$ $\to$ 8$\times$) & & 0.42s & 400K & 1.94 days \\
\hdashline
\magvit & \multirow{2}*{$8\times8\times16$} & 1.04s & 500K & 6.01 days \\
\lio (8$\times$ $\to$ 16$\times$) & & 0.23s & 500K & 2.21 days \\
\bottomrule
\end{tabular}}
\caption{Details of the per-iteration training time and total number of iterations comparing our method of progressive-growing (\lio) w.r.t to full model training (\magvit). All the model has latent dimension ($z\_dim$) = 8. The training times are computed for 17 frames of 256$\times$256 videos on H100 GPU.}
\label{tab:train_time}
\end{table}

\noindent Our progressive-growing method does not cause any training efficiency overhead. In contrast, it improves the training efficiency compared to training the full model directly for high compression. From \ref{tab:train_time}, we see that full model training from scratch (\magvit) for \eightx compression requires 2.3\textit{s/iter} while the progressive growing strategy takes 1.05\textit{s/iter}. 
Similarly, training \sixteenx model on top of \eightx model takes 0.57\textit{s/iter}, compared to 2.6\textit{s/iter} for full model training at \sixteenx compression. As a result, progressive growth allows us to train faster while achieving significantly better performance with the same number of training iterations compared to the full training approach. The main reason why our approach of progressive growing has significantly lower training time per iteration is because it does not need a discriminator to grow the model from \fourx to \eightx or from \eightx to \sixteenx temporal compression. Additionally, most of the encoder and decoder layers are frozen. For fairness of comparison, we also train the full-model training (\magvit) baseline for the same total number of iterations for a given temporal compression. Even though the full model training takes $\sim$3$\times$ more training time for \sixteenx temporal compression compared to our method \lio, our method can still achieve a lot higher quality reconstruction.

\section{Reconstruction time Number of Parameters}

\begin{table}[h]
\centering
\scalebox{0.8}{
\begin{tabular}{lcccc}
\toprule
Method & Compression & Enc. Time & Dec. Time & \# Params \\
\midrule
\magvit & \multirow{2}*{8$\times$8$\times$8} & 0.42 s & 0.47 s & 793 M \\
\lio &  & 0.61 s & 0.47 s & 794 M \\
\hdashline
\magvit & \multirow{2}*{8$\times$8$\times$16} & 0.44 s & 0.83 s & 984 M \\
\lio &  & 0.73 s & 0.83 s & 986 M \\
\bottomrule
\end{tabular}}
\caption{Details of the reconstruction time and number of parameters of our model \lio w.r.t the baseline \magvit. All the model has latent dimension ($z\_dim$) = 8. The times are computed for 17 frames of 512$\times$512 videos on A100 GPU.}
\label{tab:recon_time}
\end{table}

\begin{table}[h]
\vspace{-3mm}
\centering
\setlength{\tabcolsep}{3pt}  
\scalebox{0.75}{
\begin{tabular}{lcccccc}
\toprule
Method & Enc. (\# Params) & Enc. (Time) & Dec. (\# Params) & Dec. (Time) \\
\midrule
CogVideoX  & 92.2M & 0.14s & 123.3M & 0.32s\\
WF-VAE & 84.5M & 0.02s & 232.4M & 0.14s\\
Wan & 53.3M & 0.07s & 73.2M & 0.12s\\ 
Hunyuan & 100.3M & 0.13s & 146.1M & 0.26s\\
CV-VAE & 69M & 0.05s & 112M & 0.15s\\
\lio & 105M & 0.08s & 231M & 0.14s\\
\bottomrule
\end{tabular}
}
\caption{Details of the reconstruction time and number of parameters of encoder and decoder of our model \lio w.r.t the baselines which perform \fourx temporal compression. All the model has latent dimension ($z\_dim$) = 16. The times are computed for encoding and decoding 17 frames of 512$\times$512 resolution videos on A100 GPU.}
\label{tab:recon_time_4x}
\end{table}

\noindent For fairness of comparison, our model \lio follows the same encoder and decoder configuration to our implementation of the baseline \magvit for respective compression ratios. The difference is the progressive training strategy and the residual encoding with \texttt{AdaNorm} layers. From Table \ref{tab:recon_time}, the decoding time of our model and baseline \magvit is also the same because it follows the design. Only the encoding time of our model is ~1.5$\times$ that of baseline \magvit because we need to do two forward passes through the encoder, one with the full input and the other with temporally subsampled input (subsampled by a factor of 2). Additionally, our model \lio has almost the same number of parameters as \magvit; the only slight increase is due to the learnable parameters in \texttt{AdaNorm} blocks. Even so, for both \eightx and \sixteenx temporal compression, our model \lio has much better reconstruction quality than \magvit trained directly for \eightx or \sixteenx temporal compression. In Table \ref{tab:recon_time_4x}, we show the number of parameters for encoder and decoder along with the encoding and decoding times between our model and other baselines with \fourx temporal compression.\\
\textit{Note:} Our focus (or contribution) is \textbf{not} to have the fastest or the most efficient video tokenizer.




\include{tables/method_hyperparamaters_supp}



%% file: figures/method_supp.tex
\begin{figure*}[t!]
    \centering
    \includegraphics[width=\linewidth]{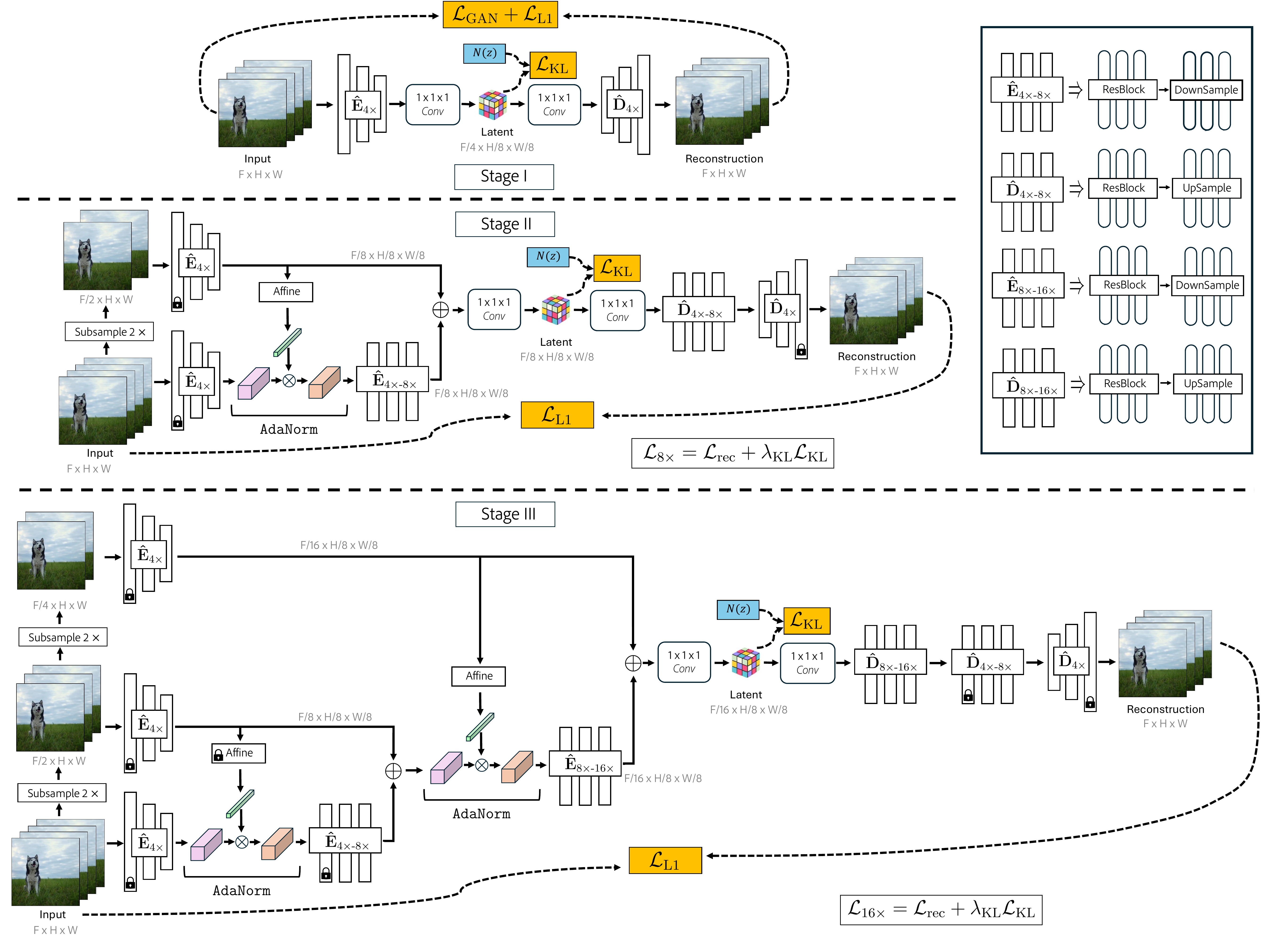}
    \caption{{\bf Methodology.} Figure shows details of our complete method of progressive growing for extending the temporal compression to \sixteenx from \fourx compression. In Stage I (top) we show a method of training our base \fourx video tokenizer. (middle) Stage II we illustrate the detailed method of growing the base \fourx model to achieve \eightx temporal compression. (bottom) Stage III we extend the \eightx temporal compression model to achieve \sixteenx temporal compression. $\mathcal{N(\text{z})}$ represent a standard normal distribution.}
    \lblfig{method_supp}
\end{figure*}

%% file: figures/overalapping.tex
\begin{figure}[t!]
    \centering
    \includegraphics[width=\linewidth]{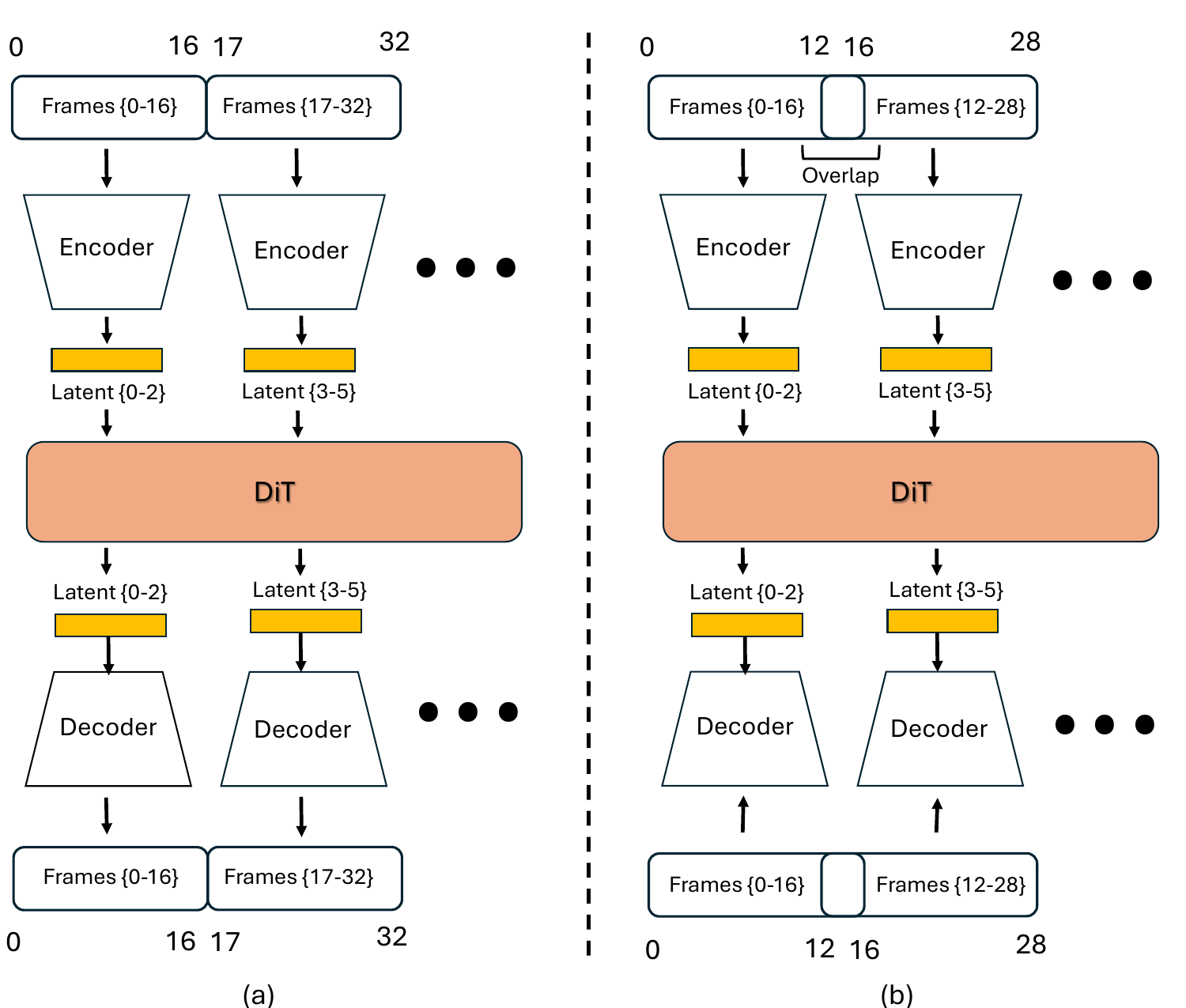}
    \vspace{-5mm}
    \caption{\camera{{\bf Text-to-Video with overlapping frames.} (left) The conventional way of generating videos, where frames are encoded into overlapping chunks. This results in jumps like artifacts across chunks in the generated video in regions of high frequency details. (right) Generating video with chunks with overlap (here of 4 frames). The overlapped regions in the output is blended in pixel-space with linear interpolated weights. This resolves the jumping artifats.}}
    \lblfig{overlapping}
\vspace{-4mm}
\end{figure}

%% file: tables/method_hyperparamaters_supp.tex
\begin{table*}[!t]
\centering
\begin{tabular}{lccc}
\hline
Encoder Config & \fourx & \eightx & \sixteenx \\
\hline
inputs & pixels & pixels & pixels\\
input size & $17 \times 256 \times 256$ & $17 \times 256 \times 256$ & $17 \times 256 \times 256$ \\
video fps & 6 & 12 & 24 \\
latent dimension & $5 \times 32 \times 32$ & $3 \times 32 \times 32$ & $2 \times 32 \times 32$ \\
\texttt{Conv}-type & \texttt{CausalConv3D} & \texttt{CausalConv3D} & \texttt{CausalConv3D} \\
base channels & 128 & 128 & 128 \\
channel multipliers & 1,2,4,6 & 1,2,4,6,6 & 1,2,4,6,6,6 \\
spatial downsampling strategy & true,true,true,false & true,true,true,false,false & true,true,true,false,false,false \\
temporal downsampling strategy & false,false,true,true & false,false,true,true,true & false,false,true,true,true,true \\
downsampling strategy & strided \texttt{Conv} & strided \texttt{Conv} & strided \texttt{Conv}\\
number of residual blocks & 2 & 2 & 2\\
z\_channels & 256 & 256 & 256 \\
z\_dim (number of channels in latent) & 8 (or 16) & 8 (or 16) & 8 (or 16) \\
\hline

Decoder Config & & & \\
\hline
outputs & pixels & pixels & pixels\\
input (latent) dimension & $5 \times 32 \times 32$ & $3 \times 32 \times 32$ & $2 \times 32 \times 32$ \\
output size & $17 \times 256 \times 256$ & $17 \times 256 \times 256$ & $17 \times 256 \times 256$ \\
\texttt{Conv}-type & \texttt{CausalConv3D} & \texttt{CausalConv3D} & \texttt{CausalConv3D} \\
base channels & 128 & 128 & 128 \\
channel multipliers & 6,4,2,1 & 6,6,4,2,1 & 6,6,6,4,2,1 \\
spatial downsampling strategy & true,true,true,false & false,true,true,true,false & false,false,true,true,true,false \\
temporal downsampling strategy & false,true,true,false & true,false,true,true,false & true,true,false,true,true,false \\
downsampling strategy & nearest + \texttt{Conv} & nearest + \texttt{Conv} & nearest + \texttt{Conv}\\
number of residual blocks & 3 & 3 & 3\\
z\_channels & 256 & 256 & 256 \\

\hline

Discriminator Config & & & \\
\hline
discriminator type & patchGAN & \multirow{7}*{None} & \multirow{7}*{None} \\
inputs & pixels &  &\\
input size & $17 \times 256 \times 256$ &  &  \\
number of layers & 3 & & \\
kernel size & $3 \times 4 \times 4$ & & \\
base channels & 64 & & \\
\texttt{Conv}-type & \texttt{Conv3D} & & \\

\hline

\end{tabular}

\caption{Details of the encoder and decoder configurations of our video tokenizer \lio, and the discriminator configuration for different temporal compression ratios \fourx, \eightx and \sixteenx. We use discriminator only for training the \fourx base model.}
\label{tab:hyperparameters}
\end{table*}